# Enhancing Mobile Face Anti-Spoofing: A Robust Framework for Diverse Attack Types under Screen Flash

Weihua Liu, Chaochao Lin, Yu Yan et al.

Beijing Institute Technology & AthenaEyesCO.,LTD.

**Abstract:** As the usage of mobile face recognition systems becomes increasingly prevalent, it is crucial to ensure robust face anti-spoofing (FAS) on mobile devices. However, existing end-to-end FAS methods, primarily reliant on handcrafted binary or pixel-wise labels, demonstrate limited robustness. The intrinsic variations among diverse types of spoof faces pose a challenge in extracting effective and universal features while accurately discriminating between live and spoof faces. This issue arises because the variations between various attack types expand the intra-class distance among spoof faces, making it difficult to learn optimal decision boundaries. To address these concerns, we propose an attack type robust face anti-spoofing framework under screen flash on mobile devices, denoted as ATR-FAS. Our framework decomposes FAS problem into two stages, which initially predicts depth maps through a mixture of type-differentiated multi-expert networks and subsequently distinguishes spoof faces based on the generated depth maps. To achieve this, we employ multiple networks to reconstruct multi-frame depth maps, and each network expertise in a specific attack type. A dual gate module (DGM) is introduced, consisting of a type gating network and a attention gating network. The DGM serves a dual purpose: recognizing attack types and generating multi-frame attention maps. The outputs of the DGM are utilized as weights to mix the result of multiple expert networks. This multi-experts mixture empowers ATR-FAS to yield spoof-differentiated depth maps, ensuring performance across varied presentation attack types. Furthermore, we innovate a differential normalization technique to transform original flash frames into differential frames. This processing step effectively mitigates the influence of ambient light and enhances the details within flash frames. This augmentation substantially contributes to the precision of depth map generation. To validate the effectiveness of our framework, we collect a large-scale dataset containing 12,660 live and spoof videos with diverse presentation attacks under dynamic flash from smartphone screens. Extensive experiments illustrate that the proposed ATR-FAS significantly outperforms existing state-of-the-art methods. The code and dataset will be available at https://github.com/Chaochao-Lin/ATR-FAS.

## 1 Introduction

Face recognition systems are prevalent on mobile devices, the application ranging from access control to mobile payment. Nonetheless, their security is still a concern both in academia and industry [1]. In response to various attacks, mobile face anti-spoofing (FAS) technology is necessary. In its early phases, FAS predominantly relied upon handcrafted features like SIFT [3], SURF [4], and HOG [5], etc. Previous studies also explored liveness cues for FAS, incorporating features like eye-blinking [6][7], nodding [8][9], and gaze tracking [10].

In recent years, deep learning approaches have gained substantial attention

[11][12][13] in the field of FAS. These methods treat FAS as a binary classification task (e.g. '0' for live and '1' for spoof faces) and adopt end-to-end training of a single neural network, as shown in the above of Fig. 1(a). According to the fact that the face surfaces of most attacks exhibit distinct depth distributions and abnormal reflections compared to live faces [1], some researches introduce depth information as intermediate auxiliary supervision within the single network for FAS [11][14]. This auxiliary supervision guides deep neural networks to capture physical spoofing clues effectively. However, the limitations of single network render these methods susceptible to different attack types, such as print, replay, and 3D masks. This susceptibility arises from the essence of deep learning, which is centers on learning data distribution [18]. Diverse attack types lead to increased the intra-class distance among. For instance, though the characteristics of print and 3D mask attacks are distinct, they are both categorized as spoofs, as shown in the above of Fig. 1(b). Without differentiation of detailed change caused by attack types, it is difficult to model the features of spoof faces while learning the optimal decision boundary. Thus, a single network cannot extract effective and universal features, while accurately distinguishing between live and spoof faces. As there are various types of spoof faces in real scene [2], this problem poses a considerable challenge for mobile FAS.

In addition, considering that obtaining depth maps requires specialized sensors, it may raise hardware costs. To explore FAS deployment on mobile devices, the integration of dynamic flash from smartphone screen coupled with a visible RGB camera (VIS-Flash) emerges as a promising pathway without additional hardware, as depicted in Fig. 1(c). Previous researches [15][16][17] illustrate the potential of VIS-Flash to simulate pseudo depth information by reflection clues. While VIS-Flash provides a possible solution for FAS deployment on mobile devices, research in the domain of FAS leveraging VIS-Flash remains relatively nascent.

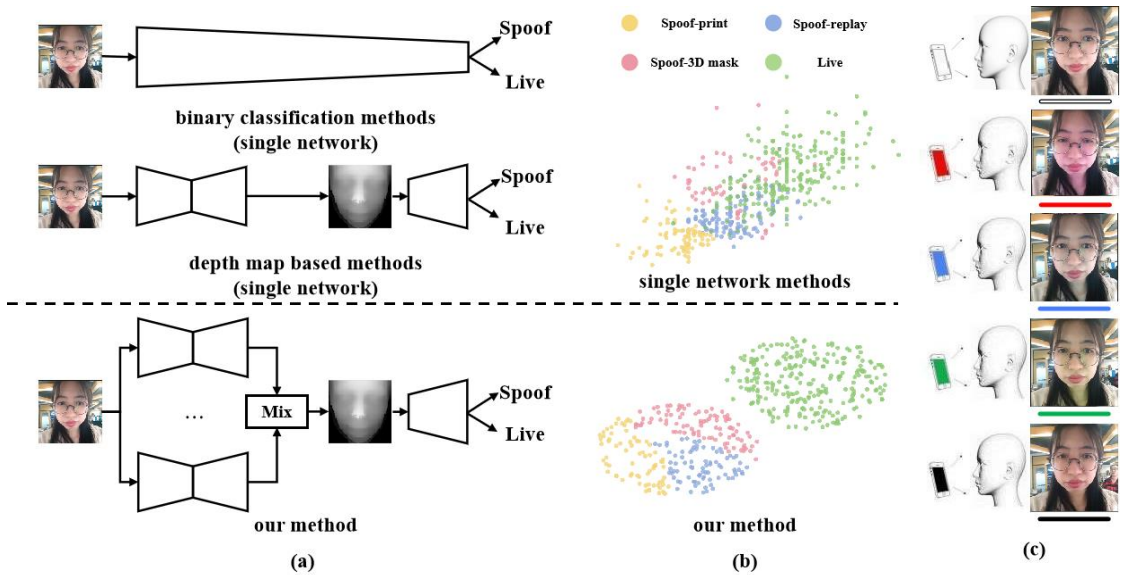

**Fig 1:** The source of inspiration for our method. (a) The framework of our method and other methods. (b) Scatter plots of features extracted by our method compared to other single network methods. (c) The example of dynamic flash and the flash frames by the smartphone.

Our framework decomposes FAS problem into two stages, which initially predicts depth maps through a mixture of type-differentiated multi-expert networks and subsequently distinguishes spoof faces based on the generated depth maps.

To address these challenges, we propose an attack type robust face anti-spoofing framework under screen flash, called ATR-FAS. Considering the imaging differences among diverse presentation attacks, our framework decomposes FAS problem into two stages, which initially predicts depth maps through a mixture of type-differentiated multi-expert networks and subsequently distinguishes spoof faces based on the generated depth maps. To achieve this, our framework involves multiple networks to reconstruct depth maps from flash frames. Each network serves as a expert in a specific attack type. By focusing exclusively on an attack type, experts networks can effectively capture the detailed features without interference from other types. In order to mix the outputs of these multi-experts and incorporate information from multiple frames, we introduce a dual gate module (DGM). The DGM comprises a type gating network and a attention gating network. The type gating network performs attack type recognition, and its recognition probability acts as a weighted operator to select and modulate the outputs of multi-expert networks. The attention gating network generates attention maps to fuse information from depth maps across frames. By multi-experts mixture, our framework reduces the distance of intra-class feature in spoof faces, thereby obtaining more spoof-differentiated depth maps to assist classification. We further design an effective differential normalization method to convert original flash frames into differential frames. This processing step utilizes the distinctive reflection of dynamic flash from smartphone screens, effectively reducing the influence of ambient light and enhancing the details within flash frames. The main contributions of this paper are summarized as follows:

1) We propose ATR-FAS, an attack type robust face anti-spoofing framework leveraging screen light. It employs type-differentiated multi-experts mixture of depth maps to enhance FAS accuracy. The mixture of type-differentiated multiple expert networks addresses the challenges of feature difference posed by attack type diversity, rendering our framework robust to various attack types.

2) The proposed ATR-FAS framework includes a dual gate module (DGM) to select and modulate the results of multiple expert networks. The type gating network and attention gating network of DGM, improve robustness against different attack types and integrates multi-frame information.

3) A differential normalization technique is introduced to convert original flash frames into differential frames. This method effectively enhance the details of flash frames.

4) To verify the effectiveness of our ATR-FAS, we collected a large-scale dataset containing 12,660 live and spoof videos with diverse presentation attacks under dynamic flash from the smartphone screen. Extensive experiments are conducted and demonstrate the effect of multi-experts mixture, and prove that ATR-FAS is superior to the state-of-the-art methods.

The remainder of this paper is organized as follows. Section 2 reviews the related works. Section 3 and Section 4 presents our ATR-FAS architecture and dataset respectively. Experimental results are reported in Section 5. Finally, we provide some concluding remarks in Section 6.

## 2 Related Works

### 2.1 Face Anti-Spoofing

Traditional FAS techniques mostly focus on handcrafted features and traditional machine learning methods such as support vector machine (SVM). Komulainen et al. [5] adopted histogram of oriented gradients (HOG) descriptors to describe context information and an SVM for detecting presentation attacks based on HOG features. Boulkenafet et al. [4] extracted speeded-up robust features (SURF) descriptions from different color spaces and fed face representation into a softmax classifier for discriminating real or spoof faces. In addition, some studies considered liveness cues for FAS. For instance, Pan et al. [6] proposed a conditional model against photograph spoofing by blink detection. These methods are effective for print attacks, but are unable to defend of replay attacks, which contains motion information.

Recently, deep learning has dominated the FAS field. Yang et al. [19] first applied convolutional neural network (CNN) and a SVM classifier to tackle FAS problem. After that, more CNN-based FAS methods are proposed [12][13]. Besides, Deb et al. [20] utilized fully convolutional neural network (FCN) to learn local discriminative cues from face images in a self-supervised manner. Ge et al. [21] combined CNN features with long short-term memory networks (LSTM) to focus on motion cues across video frames. These methods treated FAS as a binary classification problem with cross entropy loss. However, FAS models supervised by binary loss might easily overfit and is usually a black-box. Thus, some studies utilized extra depth information for auxiliary supervision. Atoum et al. [11] leveraged auxiliary supervision of pseudo depth labels to guide the multi-scale FCN, improving the FAS performance. Wang et al. [22] proposed a temporal transformer network to capture temporal variation features based on auxiliary supervision of depth maps.

**2.2 Multi-Expert Methods**

Multi-expert methods usually train several networks independently and integrate their results. Shazeer et al. [23] simplified the mixture-of-experts layer by introducing sparsity to the output of the gating function for each instance, which significantly enhances the stability of training and diminishes the computational expenses. Sam et al. [24] introduced a hierarchical strategy to train multi-experts networks, using the fine experts to predict precise results. Hu et al. [26] utilized a multi-gate mixture-of-experts [25] to address the decline in performance resulting from class-imbalance issues of graph classification. Fedus et al. [27] further enhanced the efficiency of mixture-of-experts layer by directing individual instances to a sole expert as opposed to K experts. Zhou et al. [28] employed a strategy where experts choose the top tokens instead of allocating tokens to the highest-ranked experts. The multi-expert approach has demonstrated remarkable effectiveness across diverse visual tasks, yet its application to FAS problem remains underexplored. Within the realm of multi-expert systems, determining the means to create proficient experts and appropriately assign them to test samples represents a pivotal challenge. Taking into account the interplay between expert networks, this study employs attack types as auxiliary supervision to guide the selection of expert networks instead of random gating weight.

**2.3 Attention and Gates**

The introduction of attention and gates aimed to suppress incorrect activation while directing the focus predominantly towards features related to the particular task. Fu et al. [29] employed a self-attention mechanism to fuse contextual details of objects and mitigate the influence of background noise in the context of scene segmentation tasks. Wang et al. [30] integrated a parallax-attention mechanism with a network architecture using an attention gate to achieve a global receptive field. Tang et al. [31] incorporated an attention gate within each pyramidal level, facilitating the transmission of intricate details from the top-down pathway to the bottom-up pathway. These approaches diminishes the likelihood of erroneous positives and aids in steering the attention of models towards the task-related features, thereby augmenting overall performance. We amalgamate these benefits with multi-expert mixture to produce more distinctive pseudo depth maps that aid in supporting FAS tasks.

# 3 Methodology

In this section, we introduce the attack type robust face anti-spoofing framework ATR-FAS, which which generates depth maps to assist FAS through a multi-experts mixture. The overview architecture of the ATR-FAS framework is described at first. Next, the key components are presented, including the differential normalization procedure for input flash frames, dual gate module (DGM) for attention controlling, and multi-experts mixture for the generation of discriminative depth maps. Finally, we detail the loss function of ATR-FAS for joint optimization.

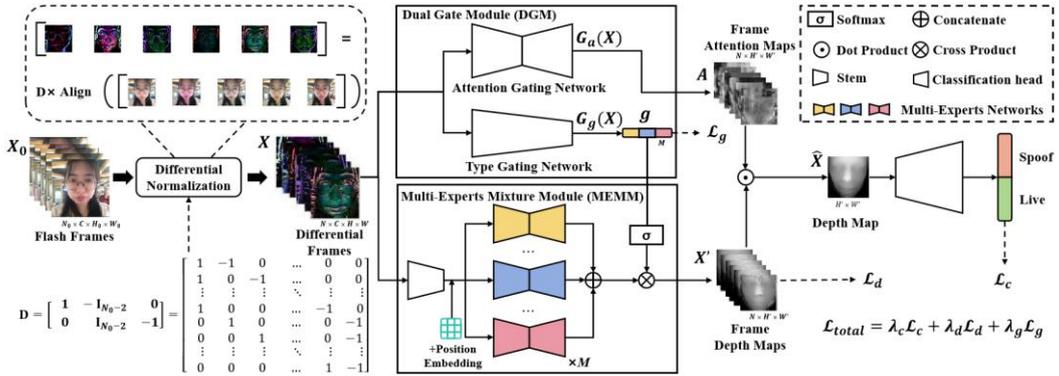

**Fig 2:** Overview of the proposed framework.

## 1.1 Overview Architecture

Inspired by the latest FAS researches, there are significant differences in depth information among live and spoof faces. However, the previous FAS methods, which use a single binary classification network for detecting presentation attacks, leading to poor attack type robustness. This is because the different attack types increase intra-class feature distances of spoof faces, which arises difficulty in a challenge of decision boundary learning. The limited modeling ability of a single network makes it difficult to extract effective and universal features while accurately distinguishing between live and spoof faces. Therefore, our framework decomposes the FAS problem into two stages, which learns the depth map features through multi-experts mixture based on attack types and differentiate spoof faces according to depth map information, as illustrated in Fig. 2.

The input flash frames of ATR-FAS are collected using dynamic flash from a smartphone screen. Let $X_0$ denotes as the input flash frames. First, we utilize differential normalization on the original flash frames $X_0$ to obtain differential frames $X$. Then, ATR-FAS employs multiple expert networks to reconstruct depth information according to the diverse attack types (e.g. print, replay and 3D masks), and each expert network corresponds to an attack type. The gating mechanisms are performed to select and modulate the result of multiple expert networks. Specifically, the differential frames flow into a dual gate module (DGM) and multi-experts mixture module (MEMM). DGM generates type weight vector $g$ and frame attention maps $A$ by a type gating network and a attention gating network respectively. By the cross product of type weight vector $g$ and multi-experts results in MEMM, frame depth maps $X'$ is gained. The dot product of frame depth maps $X'$ and frame attention maps $A$ yields final depth map $\hat{x}$. Finally, a classification head discriminates live and spoof faces based on the depth map $\hat{x}$. The ATR-FAS framework is in an end-to-end manner during the training phase. Through the multi-experts mixture, ATR-FAS can focus on the details of spoof faces under different attacks, thereby obtaining attack type robust depth maps for FAS.

**1.2 Differential Normalization**

Under dynamic flash in sequential colors with different intensity, we collect $N_0$ frames of face images. Given the original flash frames $X_0$ of ordered screen flash intensity, the transformation matrix is estimated from the facial landmarks and the affine transformation is adopted at first to align face of each frame. Then, the aligned flash frames $X'_0$ is obtained:

$$X'_0 = \text{Align}(X_0)$$

where $X_0 \in \mathbb{R}^{N_0 \times C \times H_0 \times W_0}$, $X'_0 \in \mathbb{R}^{N_0 \times C \times H \times W}$; $C$ refers to input channel size of the frames; $(H_0, W_0)$ and $(H, W)$ are the resolution of original flash frames and aligned flash frames, respectively.

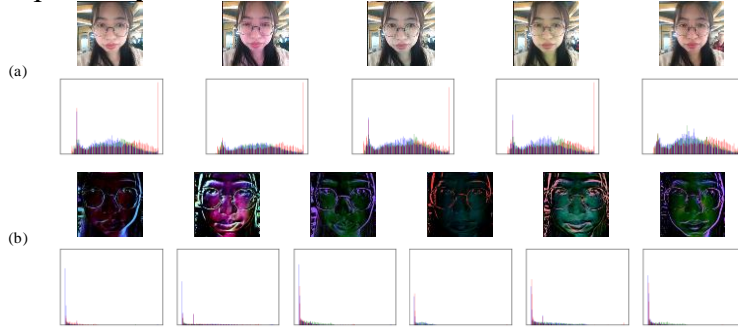

**Fig 3:** Example of frames and its histograms. (a) shows original flash frames and the corresponding histograms. (b) shows differential frames and the corresponding histograms.

Inspired by inevitable geometric relationship between depth and surface normal [37], we explore the theory for solving surface normal. According to the Lambertian reflectance model [36], the intensity at position $i$ of a face image $I$ is described theoretically as:

$$I(i) = k_a I_a(i) + k_d I_d(i) \cos \theta_i$$

where $I_a$ and $I_d$ denote as the intensity of ambient light and diffuse light respectively; $k_a$ and $k_d$ refer to the reflection coefficient of ambient light and diffuse light respectively; $\theta$ presents the angle between the direction of incident light and the surface normal. In this case, the diffuse light almost comes from screen flash.

Based on above formula, it is difficult for the deep networks to learn the functional relationship between intensity and surface normal when using the original aligned flash frames to predict the depth map. This is because different face sequences may be exposed to different ambient light. Therefore, it is necessary to eliminate the impact of ambient light. Suppose that the intensities of two face images in a flash sequence are

$I_1$ and $I_2$ respectively ($I_2 > I_1$). In the same sequence, the ambient light remains almost unchanged. For the former, the image intensities consist of $I_a$ and $I_{d,1}$, while for the latter, they have $I_a$ and $I_{d,2}$ components, as follows:

$$I_1 = k_a I_a + k_d I_{d,1} \cos\theta$$
$$I_2 = k_a I_a + k_d I_{d,2} \cos\theta$$

By subtracting $I_1$ from $I_2$, the impact by screen flash can be properly calculated. Thus, we can calculate the differential images $I_{diff}$ by:

$$I_{diff} = I_2 - I_1 = k_d [I_{d,2} - I_{d,1}] \cos\theta$$

Through differential calculation, the influence of ambient light is eliminated, retaining only the intensity change caused by screen flash. This procedure makes it easier for the deep networks to learn the robust transformation function from input intensity to surface normal and depth map. For multiple frames, there is redundancy in the differential calculation across all pairs of images. Because of the similar intensity of the screen flash between adjacent frames, the subtraction between adjacent images may cause noise. Thus, we only calculate the differential images between the face images with the highest/lowest screen flash intensity and the intermediate screen flash intensity to reduce redundancy and noise. The differential frames $X$ can be calculated by

$$X = \mathbf{D} \times X_0'$$

where $X \in \mathbb{R}^{N \times C \times H \times W}$, $\mathbf{D} \in \mathbb{R}^{N \times N_0}$ is the matrix for differential transformation, and $N = 2(N_0 - 2)$. $\mathbf{D}$ is defined as:

$$\mathbf{D} = \begin{bmatrix} \mathbf{1} & -\mathbf{I}_{N_0-2} & \mathbf{0} \\ \mathbf{0} & \mathbf{I}_{N_0-2} & -\mathbf{1} \end{bmatrix} = \begin{bmatrix} 1 & -1 & 0 & \cdots & 0 & 0 \\ 1 & 0 & -1 & \cdots & 0 & 0 \\ \vdots & \vdots & \vdots & \ddots & \vdots & \vdots \\ 1 & 0 & 0 & \cdots & -1 & 0 \\ 0 & 1 & 0 & \cdots & 0 & -1 \\ 0 & 0 & 1 & \cdots & 0 & -1 \\ \vdots & \vdots & \vdots & \ddots & \vdots & \vdots \\ 0 & 1 & 0 & \cdots & 1 & -1 \end{bmatrix}$$

where $\mathbf{I}_{N_0-2}$ denotes an identity matrix of order $N_0 - 2$, $\mathbf{1}$ represents a unit vector, and $\mathbf{0}$ is a zero vector.

**1.3 Dual Gate Module**

To address the issue of limited defense capability of FAS systems against different attack types, ATR-FAS employs multiple expert networks for each type and introduce dual gate module (DGM) to modulate the results of multi-experts. DGM is consist of type gating network and attention gating network to generate a type weight vector $g$ and frame attention maps $A$ respectively. Due to the differences between different attacks, ATR-FAS builds expert networks for each attack type. It is necessary to select the appropriate expert networks according to the type of attack. For $M$ expert networks, $g$ is an $M$-dimensional type weight vector defined as:

$$g = G_g(X)$$

where $G_g(\cdot)$ is the calculation of the type gating network. The type gating network is supervised to minimize Softmax loss function with the ground truth attack types:

$$\mathcal{L}_g = -\frac{1}{n} \sum_{i=1}^{n} \sum_{j=1}^{M} y_{i,j}^g \log \frac{e^{g_{i,j}}}{\sum_{k=1}^{M} e^{g_{i,k}}}$$

where $n$ is the batch size and $y^g$ is the ground truth attack types ($y_{i,j}^g$ denotes $i$-th sample and class $j$). Note that the ground truth of a spoof face is the $M$-dimensional one-hot code, while the ground truth of a live face is a vector with a sum of 1 and all components equal. To further improve the quality of depth images, an attention gating network imposes frame attention on the differential frames $X$. This process is denoted as:

$$A = G_a(X)$$

where $A \in \mathbb{R}^{N \times H' \times W'}$; $(H', W')$ is the resolution of frame attention maps; $G_a(\cdot)$ represents the calculation of the attention gating network. The frame attention maps focus on high-quality areas of each frame and is utilized to fuse depth maps from multiple frames into a single frame depth map, which can improve the quality of the final depth map.

**1.4 Multi-Experts Mixture**

In order to cope with the differences in presentation attacks of different types, ATR-FAS adopts a strategy of multi-experts mixture. Through multi-experts mixture module (MEMM), each expert network concentrate on one attack type and the result of multi-experts are mixed according to attack type. The multi-experts mixture suppresses the impact of increase in intra-class feature distance caused by different attack types, thereby capturing more robust depth information. This is more beneficial for learning decision boundary.

Specifically, MEMM performs spatial down-sampling on the differential frames $X$ by a stem. To retain positional information, learnable 2D position embeddings $P$ are added to the down-sampled feature maps. The resulting feature maps $X^0$ serves as input to multiple expert networks. The above process is denoted as:

$$X^0 = \text{Stem}(X) + P$$

where $\text{Stem}(\cdot)$ is the calculation of stem, $P \in \mathbb{R}^{C' \times H' \times W'}$, $X^0 \in \mathbb{R}^{N \times C' \times H' \times W'}$ and $C'$ refers to channel size of the feature maps. Then, multiple expert networks are utilized to reconstruct depth information from the feature maps. The result of each expert network is defined as:

$$X^1 = \text{Expert}^1(X^0)$$
$$X^2 = \text{Expert}^2(X^0)$$
$$\vdots$$
$$X^M = \text{Expert}^M(X^0)$$

where $X^i$ is the result of $i$-th expert network and $\text{Expert}^i(\cdot)$ refer to the calculation of $i$-th expert network, $i \in \{1, 2, \ldots, M\}$. Concatenate these results to get $\bar{\bar{X}} = [X^1, X^2, \cdots, X^M]$, and $\bar{\bar{X}} \in \mathbb{R}^{M \times N \times H' \times W'}$. The $\bar{\bar{X}}$ is weighted by the weight vector $g$ from DGM.

$$X' = \sigma(g) \times \bar{\bar{X}}$$

where $\sigma(\cdot)$ corresponds to softmax activation function and $X' \in \mathbb{R}^{N \times H' \times W'}$ represents the frame depth maps. To learn target depth information, a softmax loss function is defined.

$$\mathcal{L}_d = -\frac{1}{n} \sum_{i=1}^{n} \sum_{p \in \mathbb{Z}^2} y_i^{d(p)} \log d(p)_i$$

where $n$ is the batch size, $y^d$ is the depth label, $d(p)$ is the predicted depth value on position $p$. Finally, we fuse the frame depth maps into a single frame for FAS based on the frame attention maps $A$ from DGM. The final depth map $\hat{X} \in \mathbb{R}^{H' \times W'}$ is defined as:

$$\hat{X} = A \cdot X'$$

### 1.5 Total Loss

Based on the depth map, a classification head is implemented for discriminating live and spoof faces, as illustrated in Fig. 2. The face anti-spoofing classifier is optimized by the standard cross-entropy loss, which is denoted as:

$$\mathcal{L}_c = -\frac{1}{n}\sum_{i=1}^{n} y^c \log c + (1 - y^c)\log(1 - c)$$

where $y^c$ is the label and $c$ is the predicted probability. Integrating all objective mentioned above together, the loss function of the proposed ATR-FAS framework for face anti-spoofing is:

$$\mathcal{L}_{total} = \lambda_c \mathcal{L}_c + \lambda_d \mathcal{L}_d + \lambda_g \mathcal{L}_g$$

where $\lambda_c$, $\lambda_d$, and $\lambda_g$ is the hyper-parameters to trade-off classification loss, depth loss and gate loss in the final overall loss, respectively.

## 4 Dataset and Protocol

In this section, we first present the large-scale dataset collected under dynamic flash from the smartphone screen, and then introduce its testing protocols.

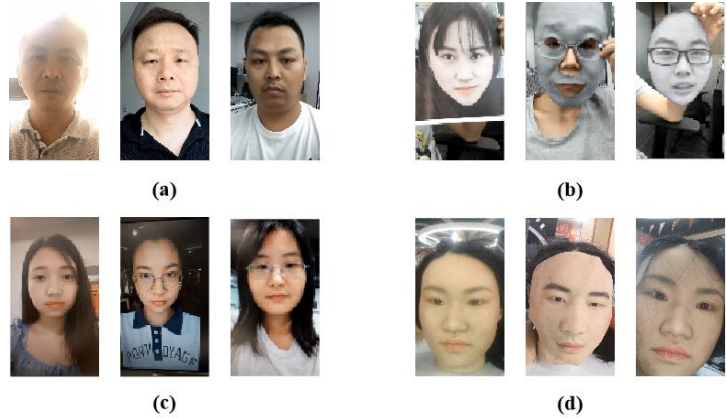

**Fig 4:** Samples of our dataset: (a) real faces, (b) print faces, (c) replay faces, (d) 3D faces.

### 4.1 Dataset

With the growing prevalence of mobile face recognition systems, the utilization of standard cameras on mobile devices for FAS has garnered substantial interest. Recent investigations suggest integrating dynamic flash from the smartphone screen to record face videos, which is proved to be efficient and lightweight [15][16][17]. Nonetheless, there presently exists no publicly accessible face image dataset with dynamic flash.

To explore the generalization of different attacks under screen flash, we collect a large-scale dataset of 12,660 videos of real and spoof faces. Real face data involves 6,620 videos from 320 subjects (roughly 20 videos per subject). For spoof faces, our collection comprises 6,040 videos covering diverse types of attacks. Specifically, it includes three parts of attacks: print, replay and 3D attacks. Among these, there are 2,120 videos pertaining to print attacks. Apart from the conventional gray, color, and high-definition papers, we introduced more demanding variations such as hole paper and cropped paper to enhance the challenge. Replay attacks involve the playback of face videos on screen media. To comprehensively cover this, we collected 2,120 videos by various screens, including mobile phone, computer, and high-definition TVs. The remaining 1,800 videos is 3D attacks, including 3D masks and 3D head models.

Within this dataset, videos are obtained from a range of mobile device brands, including Apple, Huawei, Xiaomi, and Samsung. Each video sequence lasts approximately 10 seconds, recorded at a frame rate of 30 fps. All subjects are Chinese people (167 males and 153 females). They are evenly distributed between (10, 80) years old. The youngest is 12 years old, while the oldest is 78. Considering real-world environment, data collection is performed under eight ambient light scenarios both indoor and outdoor surroundings, i.e. dark room, office, bedroom, backlight, shadow, cloudy, sunlight, and street lights. The detailed data distribution is shown in Fig. 5.

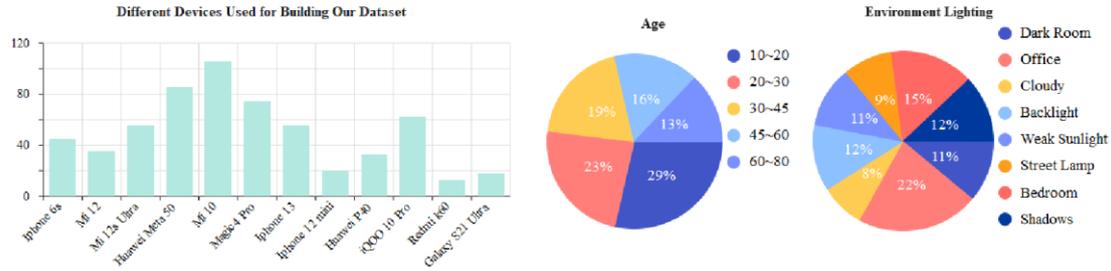

**Fig 5:** The detailed data distribution of our dataset.

### 4.2 Protocol

Our dataset adopts Intra-Dataset Intra-Type Protocol to evaluate the model robustness under domain shifts. That is, in the training and testing phase, it uses the same data set with the same attack type, but changes the acquisition conditions. In terms of performance indicators, Half Total Error Rate (HTER) and Equal Error Rate (EER) are adopted. HTER is defined as:

$$HTER = \frac{FRR + FAR}{2}$$

where FAR is the false acceptance rate and FAR is the false rejection rate. Lower HTER means better average performance on FAS, and HTER reaches its minimum when FRR=FAR, which is defined as EER.

## 5 Experiments

In this section, we conduct a series of experiments on our dataset. To begin with, we introduce the implementation details employed in the experiments. After that, we illustrate the effectiveness of the proposed framework and discuss the experimental results.

### 5.1 Implementation Details

**Face Alignment and Ground Truth Generation.** The face alignment and depth map label generation are achieved by PRNet [32]. The aligned face are with the size of 256x256, while the generated depth maps are 64x64. We normalize the live depth map in a range of [0, 1] and the spoof one is set to 0.5 at the training stage. This setting is beneficial for learning discriminative patterns for FAS task.

**Framework Details.** In our experiments, we set up three expert networks for print, replay, and 3D mask attacks, respectively. The three expert networks are ResUNet [38] with the same structure. The stem before the expert network is composed of double 3x3 convolution layers, performing initial downsampling. The attention gating network consists of two layers of 3x3 convolution layers and a simplified version of ResUNet (only contains two downsamples). The type gating network contains three 3x3 convolution layers and two linear layers, and outputs a 3-class probability. The classification head consists of three 3x3 convolutions and one linear layer. The detailed framework structure is as shown in Table 1.

**Training Setting and Hyperparameter.** We use Pytorch to implement our method. At the training stage, Adam optimizer are adopted with batchsize 4. The initial learning rate are set to 1e-4 with exponential decay. To ensure efficiency and accuracy, we extract $N_0 = 5$ flash frames from the video. Therefore, the differential matrix of differential normalization is:

$$\mathbf{D} = \begin{bmatrix} 1 & -1 & 0 & 0 & 0 \\ 1 & 0 & -1 & 0 & 0 \\ 1 & 0 & 0 & -1 & 0 \\ 0 & 1 & 0 & 0 & -1 \\ 0 & 0 & 1 & 0 & -1 \\ 0 & 0 & 0 & 1 & -1 \end{bmatrix}$$

## 5.2 Comparison

Our approach achieves a EER of 1.01% and a HTER of 1.29% for FAS task. We compare these metric of the proposed FAS method with those of other methods, including SURF [11], FASNet [33], Auxiliary Depth CNN [14], Deep LBP [34], AG [15], BASN [35] and TTN [22]. The same training and validation datasets are used for all the methods. The comparison results are shown in Table 2. These values demonstrate that the proposed approach stably outperforms comparison methods at EER and HTRR. The corresponding increases on the HTER are 13.09%, 6.19%, 4.07%, 5.94%, 2.94%, 1.23%, and 1.06% compared with SURF, FASNet, Auxiliary Depth CNN, Deep LBP, AG, BASN and TTN, respectively. And for EER, our method is 3.51%, 4.20%, 3.74%, 4.31%, 2.51%, 1.32%, and 1.02% lower than SURF, FASNet, Auxiliary Depth CNN, Deep LBP, AG, BASN and TTN, respectively. Examples of depth maps are shown in Fig. 6 compared with the best two method, BASN and TTN. This figure illustrates our method can generate more discriminative depth maps to assist in FAS. Specifically, our method generates a more three-dimensional depth map of the active face compared to BASN and TTN, while the depth map of the attacking face is smoother. The findings demonstrate that our approach achieves better FAS result attribute to distinguishable depth map generation by multi-expert mixture.

Table 2: The comparison with FAS methods, including SURF [11], FASNet [33], Auxiliary Depth CNN [14], Deep LBP [34], AG [15], BASN [35] and TTN [22].

| method | HTER (%) | EER (%) |
|---|---|---|
| SURF [11] | 14.38 | 4.52 |
| FASNet [33] | 7.48 | 5.21 |
| Auxiliary Depth CNN [14] | 5.36 | 4.75 |
| Deep LBP [34] | 7.23 | 5.32 |
| AG [15] | 4.23 | 3.52 |
| BASN [35] | 2.52 | 2.33 |
| TTN [22] | 2.35 | 2.03 |
| ATR-FAS (ours) | 1.29 | 1.01 |

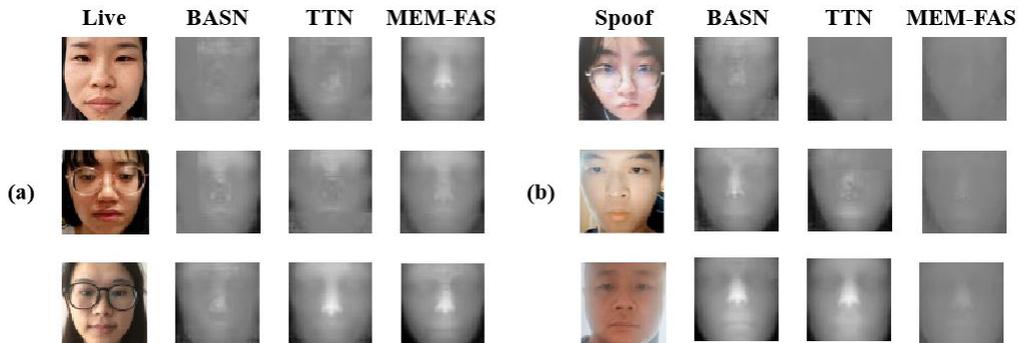

**Fig 6:** The comparison of auxiliary depth map generation with best two methods, BASN [35] and TTN [22].

### 5.3 Ablation Study

Three new ideas in this paper are MEMM, DGM and differential normalization. We study the effectiveness of each of the three contributions in this section. MEMM and DGM jointly implement the entire multi expert mixing process. Accordingly, we consider two choices for MEMM and five ablated choices for DGM. Two choices for MEMM are without MEMM (denoted as w/o MEMM) and with MEMM (denoted as MEMM). The role of DGM is to serve as a basis for selecting and modulating the output of multi-expert networks. Thus, the five ablated choices for DGM are as follows: with averaging (denoted as Avg), with sum operator (denoted as Sum), with concatenation (denoted as Cat), with attention network (denoted as ATT), with random gate network (denoted as RG), with random gate network (denoted as RG) and with type gate network (denoted as TG), with full DGM (denoted as DGM). By selecting the implementation from the above choices, we have 9 different configurations. We conduct ten-fold cross-validation experiments for each configuration. The corresponding mean performance metrics for each configuration are listed in Table 3.

### 5.3.1 The Effectiveness of Multi-Expert Mixture

Compared with the configuration without MEMM, the embedding of the multi-expert networks contributes obvious improvements in the HTER and EER as illustrated in Table 2, regardless of which method of mixture is used. To further verify the role of multi-expert mixture, we compare the FAS results various mixture in Table 2. Regardless of the choice selected for MEMM, the HTER and EER from our frame attention method is always better than that without frame attention weighting. The corresponding EER of decrease are 0.80% (w/o MEMM-ATT compared with w/o MEMM), 0.53 % (MEMM-ATT compared with MEMM-Avg), 0.28% (MEMM-RG-ATT compared with MEMM-RG), and 0.27% (MEMM-DGM compared with MEMM-TG). Compared with direct calculation (Avg, Sum, Cat), both common random gate control and our gate network according to attack type supervision increase the performance of FAS, from (2.69%/2.48%, 2.73%/2.54%, 2.38%2.32%) to 1.92%/1.84% and 1.56%/1.45%, respectively, It also shows that our gate network with attack type supervision surpasses the common gate control. These phenomenon show that the DGM method is effective and behaves better and stabler than the traditional mixing method.

Table 3: The ablation study result of multi-expert mixture.

| Setting | HTER (%) | EER (%) |
|---|---|---|
| w/o MEMM | 4.03±0.86 | 4.21±0.81 |
| MEMM-Avg | 2.69±0.47 | 2.48±0.41 |
| MEMM-Sum | 2.73±0.42 | 2.54±0.37 |
| MEMM-Cat | 2.38±0.40 | 2.32±0.36 |
| w/o MEMM-ATT | 3.23±0.62 | 3.14±0.63 |
| MEMM-ATT | 2.16±0.39 | 1.94±0.35 |
| MEMM-RG | 1.92±0.31 | 1.84±0.28 |
| MEMM-RG-ATT | 1.64±0.28 | 1.58±0.35 |
| MEMM-TG | 1.56±0.27 | 1.45±0.20 |
| MEMM-DGM (full model) | **1.29±0.33** | **1.01±0.23** |

**5.3.2 The Effectiveness of Differential Normalization**

In our implementation of ATR-FAS, before mixing the result of multi-expert networks, we perform differential normalization. In this experiment, we train ablated ATR-FAS without differential normalization. We denote ATR-FAS trained without differential normalization as w/o DN and denote the trained full ATR-FAS as DN. The corresponding experimental result is shown in Table 4. Without differential normalization, the HTER and EER decreases significantly from 2.32% to 1.29% and 2.26% to 1.01% respectively, which reveals the importance of differential normalization for ATR-FAS. In addition, we also compared our method with the method provided in aurora guard [15], which calculates normal cues by the difference between consecutive frames. As a result, we find that our method can significantly improve the performance. This is because directly subtracting continuous frames can cause significant noise, but our method can solve this problem by the differential matrix.

Table 4: The ablation study result of differential normalization.

| Metric | w/o DN | DN | Aurora guard [15] |
|---|---|---|---|
| HTER (%) | 2.32±0.57 | **1.29±0.33** | 2.05±0.36 |
| EER (%) | 2.26±0.46 | **1.01±0.23** | 1.93±0.32 |

**5.3.3 The Impact of Hyperparameter $N_0$**

In this ablation experiment, we aim to verify the impact of hyperparameter $N_0$. $N_0$ is set to 5 by default in this work. Here, we set $N_0$ to other values from 3 to 8. According to the procedure of differential normalization, the larger $N_0$ is, the lower HTER and EER of ATR-FAS due to the fusion of more flash frames. Fig. 7 shows how $N_0$ affects the performance of the ATR-FAS. When $N_0 = 8$, the ATR-FAS achieves the best performance with the lowest HTER and EER, however, it costs most time to inference. Smaller $N_0$ decreases the performance of ATR-FAS. But, it is worth noting that when $N_0$ increase from 5 to 8, the HTER and EER of ATR-FAS does not significantly decrease with increasing time cost. Therefore, we chose the most efficient $N_0$=5.

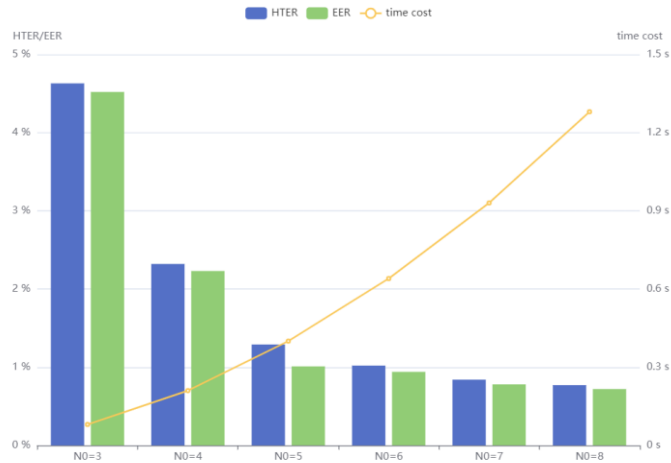

**Fig 7:** Time cost, HTER and EER of ATR-FAS with different $N_0$.

# 6 Conclusion

In this paper, we present the ATR-FAS framework for face anti-spoofing. Our method tackles the limitation of existing FAS methods and addresses the challenge posed by diverse presentation attack types. A novel multi-expert mixture architecture is designed to against the limitation of existing FAS methods caused by different types of presentation attacks. Selecting and modulating the results of multiple expert networks through weighting operators for different attack types leads to more reliable depth information, thereby improving the performance of FAS. To enhance the discriminative features between real and spoof faces, ATR-FAS adopts differential normalization to process the original flash frames. Furthermore, we collect a large-scale dataset from various mobile devices with dynamic flash. Extensive experiments on challenging FAS datasets show that ATR-FAS is superior to the state-of-the-art methods.